\documentclass{bmvc2k}

\usepackage{amsmath}
\usepackage{xcolor}
\usepackage{amssymb}
\usepackage{booktabs}
\usepackage{multirow}
\usepackage{graphicx}
\usepackage{tabularx}
\usepackage{array}
\usepackage{pifont}
\usepackage{comment}

\usepackage[table]{xcolor}
\definecolor{oursblue}{RGB}{220,245,250}
\definecolor{secondblue}{RGB}{238,250,252}

\title{BasketEvent: Understanding Who Did What and When in Basketball Videos}

\addauthor{Yu Zhang}{zhangyu2021@sjtu.edu.cn}{1}
\addauthor{Jiayuan Rao}{jy_rao@sjtu.edu.cn}{1}
\addauthor{Haoning Wu}{haoningwu3639@gmail.com}{1}
\addauthor{Weidi Xie}{weidi@sjtu.edu.cn}{1}

\addinstitution{
School of Artificial Intelligence, Shanghai Jiao Tong University\\
Shanghai, China
}

\runninghead{Zhang et al.}{BasketEvent}

\usepackage{amsmath}
\usepackage{amssymb}
\usepackage{booktabs}
\usepackage{multirow}
\usepackage{graphicx}

\newcommand{\dataset}{BasketEvent}
\newcommand{\model}{PlayNet}

\begin{document}

\maketitle

\begin{abstract}
Comprehensive basketball video understanding requires resolving not only \emph{what} event occurs, but also \emph{who} is responsible and \emph{when} the key evidence appears. 
However, existing methods typically treat spatial perception and semantic recognition as isolated tasks, failing to ground events to individual players or pinpoint their temporal boundaries within complex collective dynamics. 
To bridge this gap, we introduce \textbf{\dataset}, a player-centric basketball event understanding dataset curated from real NBA broadcasts. 
In \dataset, event labels are grounded to the responsible players, and a manually annotated subset of 1,000 samples with precise event intervals is provided to evaluate temporal evidence localization. 
Based on this data, we propose \textbf{\model}, a player-centric reasoning framework that maps basketball videos to player-level event predictions with temporal evidence. 
Concretely, \model\, tracks key entities, associates player identities, and reasons about events by modeling player-player, player-ball, and global court interactions, while aggregating sparse temporal evidence via gated pooling. 
Extensive experiments demonstrate that \model\, significantly outperforms representative video-level and crop-based baselines, proving the superiority of player-centric modeling for fine-grained sports video understanding. 
Our data, code, and models will be made publicly available.

\end{abstract}

\section{Introduction}
\label{sec:intro}

\begin{figure*}[t]
    \centering
    \includegraphics[width=\textwidth]{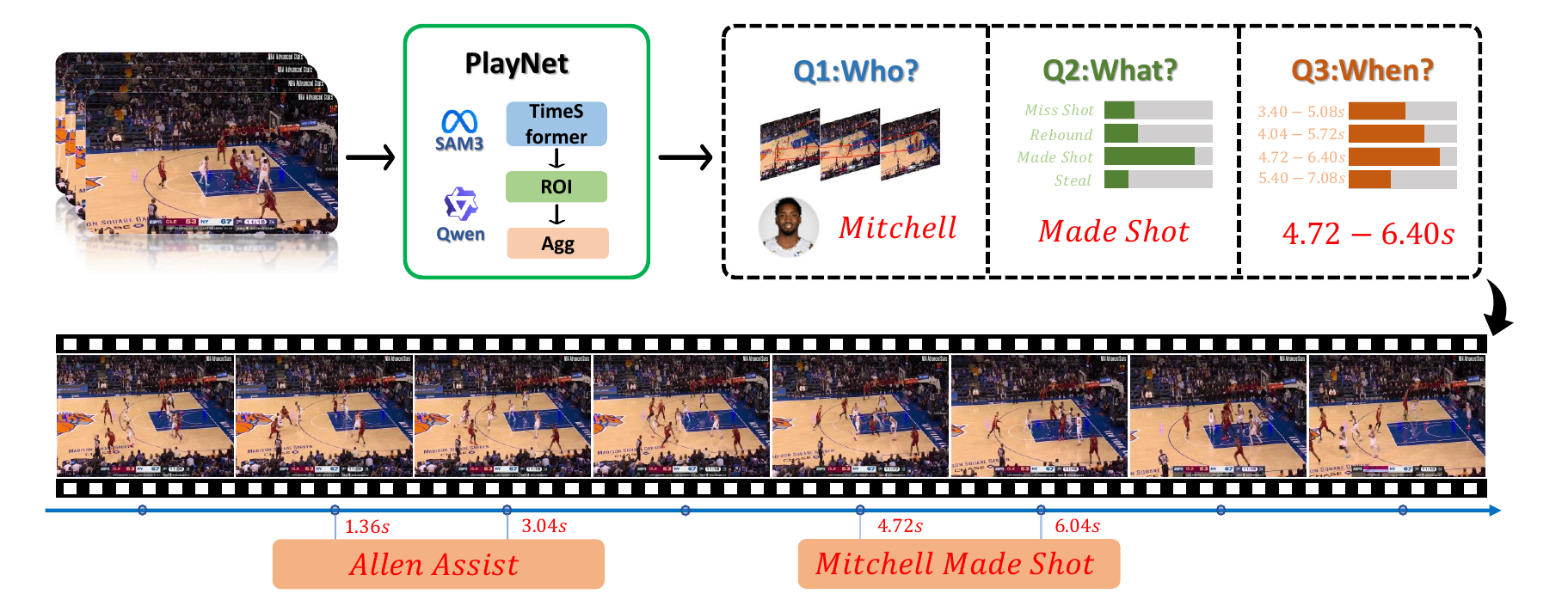}
    \vspace{-12pt}
    \caption{
        \textbf{Overview of the proposed task and pipeline.}
        Given an input basketball video, our framework answers three player-centric questions: {\em who} performs the event, {\em what} event occurs, and {\em when} the key evidence appears.
    }
    \label{fig:overview}
    \vspace{-12pt}
\end{figure*}

Sports videos provide a rich setting for studying fine-grained video understanding. 
They capture complex human motion, dense multi-agent interactions, object dynamics, spatial structure, and rapid decision-making within a shared competitive environment. 
Unlike many everyday action-recognition benchmarks, where a short clip can often be described by a single actor performing a coarse action, sports events emerge from the coordinated behavior of multiple players, the motion of the ball, the geometry of the playing field, and the temporal evolution of tactics and intent~\cite{soccernet, soccernetv2, SoccerNetv3, CameraCalibration, rao2024matchtimeautomaticsoccergame, rao2025unisoccer, rao2025socceragent, yang2025soccermaster, SoccerNet-GSR, SoccerNet-Camera, held2023vars}. 

Basketball is a particularly challenging instance of this problem. 
In a typical broadcast clip, ten visually similar players move rapidly across the court, frequently occlude one another, exchange possession within fractions of a second, and appear under fast camera motion and changing viewpoints. 
Moreover, the decisive evidence for an event is often subtle and brief: a {\em shot} may be determined by the release and ball trajectory; a {\em steal} by a momentary hand-ball interaction; a {\em foul} by transient body contact; and a {\em rebound} by spatial position and possession change. 
Thus, recognizing that an event occurs is only a partial solution. 
A basketball understanding system must jointly answer three coupled questions: \emph{who} is responsible, \emph{what} event occurs, and \emph{when} the decisive evidence appears.

Existing basketball video understanding remains fragmented. 
One line of work studies spatial perception, including player detection, tracking, and identity association in broadcast videos~\cite{sportsmot, NBAidentity}; another focuses on semantic recognition, such as group activity recognition, event classification, action localization, and identity-aware captioning~\cite{social, li2021multisports, finesports}. 
Although these tasks provide important foundations, they are typically formulated independently: perception-centric methods recover where players are, but not what role each player plays in an event, while recognition-centric methods infer what happened, but often only at the clip level. 
Consequently, existing benchmarks~\cite{VCNBA, XI2025126906} largely evaluate whether models can recognize basketball events, rather than whether they can explain them through the responsible players and decisive moments.


In this work, we introduce a player-centric formulation for basketball video understanding: given a broadcast clip, the goal is to determine \emph{who did what and when}. 
This formulation shifts the unit of understanding from the video clip to the player-event instance, explicitly grounding event labels to responsible players and localizing the temporal intervals in which the decisive evidence occurs.

To support this task, we construct \textbf{\dataset}, a player-centric basketball event understanding dataset curated from real NBA broadcasts. We leverage the NBA API~\cite{nba_api} to collect play-by-play records and retrieve the corresponding event-centered broadcast clips from the Internet. 
We then design an automatic annotation pipeline that transforms video-level play metadata into player-level event annotations by associating each event with its responsible athlete. 
This yields videos annotated with player-grounded event labels, enabling models to infer not only \emph{what} happened but also \emph{who} caused it. 
To evaluate temporal grounding, we further manually annotate \textbf{1K} player-level samples with precise event intervals.

Building on the dataset, we propose \textbf{\model}, a player-centric reasoning framework for temporally grounded basketball event understanding. 
Given an input broadcast video, \model\, first obtains player and ball trajectories across frames and associates player tracks with identities. 
It then extracts trajectory-guided visual tokens, incorporates global court context, and models player-player as well as player-ball relations around each candidate athlete. 
Finally, a gated evidence aggregation module pools sparse but discriminative temporal cues to jointly predict player-level events and localize their supporting intervals.

Extensive experiments show that player-centric reasoning provides a more faithful representation of basketball gameplay than conventional video-level or crop-based event recognition. 
By grounding events in both players and time, \model\, enables more fine-grained, interpretable, and temporally precise basketball video understanding.

In summary, our contributions are:
(i) we introduce a player-centric formulation for basketball video understanding that jointly addresses \emph{who} is responsible, \emph{what} event occurs, and \emph{when} the decisive evidence appears;
(ii) we construct \textbf{\dataset}, a player-centric basketball dataset from real NBA broadcasts, with event labels grounded to responsible players and a manually annotated subset with 1K samples for temporal localization evaluation;
(iii) we propose \textbf{\model}, a player-centric reasoning framework that combines trajectory-guided visual representations, global court context, and multi-agent interaction modeling for temporally grounded player-level event prediction;
(iv) we conduct extensive experiments showing that player-centric modeling yields more detailed and interpretable basketball video understanding than conventional video-level or crop-based event recognition.

\section{Related Work}
\label{sec:related}

\noindent \textbf{Sports video datasets} have evolved from coarse video-level recognition to fine-grained spatio-temporal and semantic understanding.
In soccer, SoccerNet and its extensions~\cite{soccernet, soccernetv2, matchtime} lead the way in long-form event spotting, temporal localization, and commentary generation.
For individual sports, FineGym~\cite{finegym} and FineDiving~\cite{finediving} focus on procedure-aware understanding with fine-grained temporal action structures.
In basketball, existing datasets~\cite{social, multisports, finesports, NBAidentity} primarily target group activity recognition, spatio-temporal action localization, and identity-aware captioning.
However, these benchmarks typically provide video-level labels but still lack fine-grained, player-level event annotations.
Our work bridges this gap by constructing a player-centric basketball benchmark that grounds complex events to the responsible players, enabling fine-grained, player-level event understanding.

\vspace{3pt}
\noindent \textbf{Sports video understanding} has advanced significantly across diverse tasks such as event spotting~\cite{soccernet, context-aware, soccernetv2, Temporally-aware}, action quality assessment~\cite{actionqualityassessment, Group-aware}, player tracking~\cite{sportsmot, Soccernet-tracking, teamtrack}, tactical analysis~\cite{tacticai, gentac}, and commentary generation~\cite{SoccerNet-caption, goal, matchtime, descriptive}.
However, most existing methods primarily focus on identifying what event occurs and when it happens, while the underlying player-centric structure of team sports remains underexplored.
This limitation is particularly acute in basketball, where understanding an event inherently requires identifying the responsible player and reasoning about his interactions with teammates, opponents, and the ball.
To address this, our work focuses on player-centric basketball event understanding, aiming to simultaneously recognize the event category, identify the responsible player, and localize the supporting temporal evidence.

\vspace{3pt}
\noindent \textbf{Player-centric interaction modeling} has emerged as a crucial paradigm in multi-person action understanding, showing that recognizing an actor's behavior requires going beyond isolated appearance cues. 
While early group activity recognition methods aggregate individual representations~\cite{hierarchical, bagautdinov2017social} or learns relations among actors through relational representations or actor relation graphs~\cite{hra, prl, acg}, recent approaches incorporate broader scene context and higher-order relations~\cite{actiontransformer, acrn, acar}. 
Furthermore, object-centric video models show that explicit object representations are critical for resolving human-object interactions~(HOI)~\cite{orvit, li2025simultaneous}.
This is particularly relevant to basketball, where key events~({\em e.g.}, assists, steals, and rebounds) are inherently defined by the coordinated dynamics between the target player, surrounding players, and the ball. 
To leverage these insights, our method constructs explicit player and ball tokens, progressively modeling global-player interaction, player-ball interaction, and temporal evidence for player-centric event recognition.

\begin{figure*}[t]
    \centering
    \includegraphics[width=\textwidth]{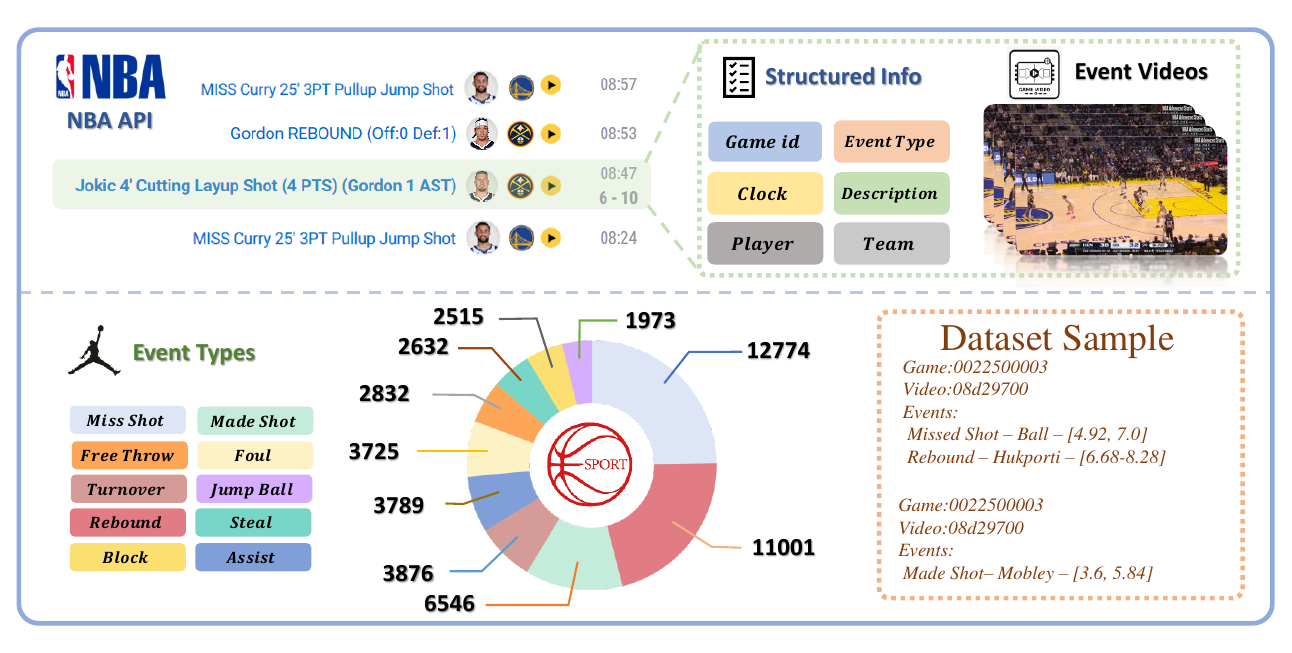}
    \vspace{-15pt}
    \caption{
    \textbf{Overview of \dataset.} 
    We collect event-centered basketball videos from NBA play-by-play records, extract and normalize structured event data into 10 representative classes, and construct player-grounded samples with event labels and temporal annotations.
    }
    \label{fig:dataset_overview}
    \vspace{-8pt}
\end{figure*}

\section{Dataset}
\label{sec:dataset}

This section describes the construction of \textbf{\dataset}. 
Sec.~\ref{subsec:dataset_construction} details how we collect broadcast videos and transform official play-by-play records into player-grounded event annotations. 
Sec.~\ref{subsec:statistics_split} summarizes the resulting dataset statistics and compares \dataset\, with existing benchmarks. 
Sec.~\ref{subsec:temporal_subset} introduces a temporally annotated subset, built from the test split, for evaluating evidence localization. 
Together, these components make \dataset\, a benchmark for fine-grained basketball video understanding in which events are not only recognized, but also grounded to the responsible players.

\subsection{Dataset Construction}
\label{subsec:dataset_construction}

As shown in Fig.~\ref{fig:dataset_overview}, \dataset\, is constructed through a scalable annotation pipeline that aligns official NBA play-by-play records with broadcast video and converts game logs into player-level events. 
The pipeline consists of two stages: event-centered video collection and player-grounded label construction.

\vspace{3pt} \noindent \textbf{Data collection.}
We collect official play-by-play records from real NBA games using the NBA API~\cite{nba_api}. Each record provides structured game metadata, including the game ID, game clock, event type, textual description, and participating players. 
Using the game ID and game clock as temporal anchors, we locate the corresponding broadcast footage and crop it into an event-centered video clip. 
This process produces an initial set of video-event pairs, where each clip is aligned with a specific play in the official game log.

\vspace{3pt} \noindent\textbf{Player-grounded event construction.}
The play-by-play data contain complementary sources of supervision: structured event fields and natural-language descriptions. 
We first extract six event categories directly from the structured fields: \textit{Missed Shot}, \textit{Made Shot}, \textit{Free Throw}, \textit{Foul}, \textit{Turnover}, and \textit{Rebound}. 
These events are explicitly recorded by the official game log and are associated with an event player, which provides a direct mapping from event type to corresponding athlete.

Structured fields alone, however, do not fully capture interaction-dependent basketball events. For example, assists, blocks, steals, and jump balls are often expressed through the textual description of a play rather than as the primary event type. 
We therefore parse the descriptions with deterministic linguistic templates to recover four additional categories: \textit{Assist}, \textit{Block}, \textit{Steal}, and \textit{Jump Ball}. 
For instance, the description ``{\em Jump Ball Holmgren vs.~Adams: Tip to Thompson}'' identifies a jump-ball event involving Holmgren and Adams; we accordingly assign the \textit{Jump Ball} label to both participating players. 
Similarly, assisted shots, blocked attempts, and stolen possessions are parsed to identify the corresponding assisting, blocking, or stealing player.
After normalization, \dataset\, contains a set $\mathcal{Y}$ of 10 player-grounded event categories: \textit{Missed Shot}, \textit{Made Shot}, \textit{Free Throw}, \textit{Foul}, \textit{Turnover}, \textit{Rebound}, \textit{Assist}, \textit{Block}, \textit{Steal}, and \textit{Jump Ball}, covering the major semantic units needed to describe basketball plays, together with a \text{Background} class for players who are not involved in any of the target events within the corresponding clip. 
As a result, all the event categories are in $\mathcal{Y}^{+}=\mathcal{Y}\cup\{\text{Background}\}$.

Finally, because basketball events are temporally coupled, multiple player-event annotations may correspond to the same visual segment. 
A made field goal may co-occur with an assist; a missed shot may be immediately followed by a rebound; and a turnover may be paired with a steal. 
Rather than treating these as independent clips, we merge overlapping segments and attach all valid player-event pairs to the unified video. 
The resulting samples are therefore multi-event and player-grounded, providing supervision that reflects the compositional nature of real basketball plays.

\begin{table*}[t]
    \centering
    \small
    \renewcommand{\arraystretch}{1.10}
    \caption{
    \textbf{Dataset comparison.}
    We compare \dataset\, with representative datasets from soccer, basketball, and gymnastics in terms of dataset scale and annotation granularity. Here, \textbf{\#Videos} denotes the number of video clips or video segments contained in each dataset, while \textbf{\#Samples} denotes the number of annotated instances provided by the dataset. Depending on the dataset, a sample may correspond to an event annotation, an action instance, a caption, or a player-level label. In our player-centric basketball event recognition task, each sample refers to one annotated event instance grounded to a target player. ``--'' indicates that the statistic is not explicitly reported.
    }
    \label{tab:dataset_comparison}
    \vspace{5pt}
    \resizebox{0.98\textwidth}{!}{
    \begin{tabular}{lccccc}
    \toprule
    \textbf{Dataset} 
    & \textbf{Domain} 
    & \textbf{\#Videos} 
    & \textbf{Duration} 
    & \textbf{\#Samples} 
    & \textbf{Granularity} \\
    \midrule

    SoccerNet~\cite{soccernet}
    & Soccer & 1k & 764h & 6k & Video \\

    SoccerNet-v2~\cite{soccernetv2}
    & Soccer & 1k & 764h & 300k & Video \\

    SoccerNet-Caption~\cite{SoccerNet-caption}
    & Soccer & 0.9k & 715.9h & 36k & Video \\

    MatchTime~\cite{matchtime}
    & Soccer & 0.9k & 715.9h & 36k & Video \\

    SoccerReplay-1988~\cite{rao2025unisoccer}
    & Soccer & 150k & 3,323h & 150k & Video \\
    
    MultiSports\textsuperscript{S}~\cite{multisports}
    & Soccer & 0.8k & 5.02h & 12k & Player \\
    
    \midrule

    NBA~\cite{social}
    & Basketball & 9k & -- & 9k & Video \\

    MultiSports\textsuperscript{B}~\cite{multisports}
    & Basketball & 0.8k & 4.28h & 9k & Player \\

    FineSports~\cite{finesports}
    & Basketball & 10k & 32.6h & 16k & Player \\

    NBA-Identity~\cite{NBAidentity}
    & Basketball & 9k & 8.9h & 9k & Video \\

    BH-Commentary~\cite{descriptive}
    & Basketball & 4.3k & 10.1h & 4.3k & Video \\

    \midrule

    FineGym~\cite{finegym}
    & Gymnastics & -- & 708h & 32k & Action \\

    MultiSports\textsuperscript{A}~\cite{multisports}
    & Gymnastics & 0.8k & 6.82h & 8.7k & Player \\

    \midrule
    
    \rowcolor{oursblue}
    \textbf{\dataset~(Ours)}
    & \textbf{Basketball}
    & \textbf{35k}
    & \textbf{90.6h}
    & \textbf{51k}
    & \textbf{Player} \\

    \bottomrule
    \end{tabular}
    }
    \vspace{-3pt}
\end{table*}

\subsection{Data Statistics}
\label{subsec:statistics_split}

We summarize the scale, annotation structure, and data split of \textbf{\dataset}, and compare it with representative basketball and sports video benchmarks to clarify its position among existing resources.

\vspace{3pt}
\noindent \textbf{Comparison with existing datasets.}
As shown in Tab.~\ref{tab:dataset_comparison}, \dataset\, contains \textbf{35k} broadcast videos~(\textbf{90.6} hours), and provides \textbf{51k} player-level event samples. Beyond scale, the key distinction of \dataset\, lies in its supervision granularity. Most existing basketball or sports video datasets focus on either video-level semantic labels or isolated perception annotations, such as player detection, tracking, or identity association. In contrast, \dataset\, unifies event semantics with player grounding by providing player-level event labels together with player and ball trajectories. This design supports a more demanding evaluation protocol: models must recognize what happened, associate the event with the responsible athlete, and reason over player-player and player-ball interactions within the same play.

\vspace{3pt}
\noindent \textbf{Data split.}
To avoid near-duplicate clips across splits and to evaluate generalization to unseen games, we partition \dataset\, at the game level rather than the clip level. The training split contains \textbf{189} games, \textbf{30,725} videos, and \textbf{45,683} player-level samples, while the test split contains \textbf{37} games, \textbf{4,206} videos, and \textbf{5,980} player-level samples. In total, \dataset\, spans \textbf{226} games, \textbf{34,931} videos, and \textbf{51,663} player-level samples.

\subsection{Temporal Evaluation Subset}
\label{subsec:temporal_subset}

Player-grounded event labels specify \emph{who} performs an event, but they do not by themselves indicate \emph{when} the visual evidence occurs. To evaluate this temporal grounding ability, we construct a manually annotated temporal evaluation subset from the test split of \dataset. 
This setting is important for basketball, where many events are defined by short-lived cues: a shooting release, a hand-ball deflection, body contact, or the instant of possession change may last only a few frames. 
However, existing basketball benchmarks rarely provide temporal boundaries for such fine-grained event evidence within broadcast clips.

To ensure that temporal localization is evaluated under the same generalization setting as event recognition, we sample all temporally annotated examples exclusively from unseen test games. 
For each of the ten event categories, we randomly select \textbf{100} player-level samples from the test split and manually annotate the start and end timestamps of the target event. 
This results in \textbf{1,000} temporally annotated player-level samples. 
The subset enables a joint evaluation of whether a model can identify the responsible player, recognize the event category, and localize the temporal interval containing the supporting visual evidence.

\section{Method}
\label{sec:method}

Here, we start by presenting \model, a player-centric framework for temporally grounded basketball event understanding. Sec.~\ref{subsec:problem_formulation} first formalizes the task and defines the player-level output space. Sec.~\ref{subsec:trajectory_tracking_player_recognition}--Sec.~\ref{subsec:feature_aggregation} then describe the main components of \model: player and ball grounding, trajectory-guided visual representation, and interaction-aware evidence aggregation. 
Finally, Sec.~\ref{subsec:output_training} presents the  training objective, and inference procedure.

\subsection{Problem Formulation}
\label{subsec:problem_formulation}
Given a broadcast basketball video, our goal is to infer structured \emph{what-who-when} event tuples: \emph{what} event occurs, \emph{who} is responsible for the event, and \emph{when} the supporting visual evidence appears. 
Formally, let $\mathcal{V}=\{I_1, I_2, \dots, I_T\}$ denote an input video with $T$ frames, we formulate basketball event understanding as a player-centric structured prediction problem. For each video $V$, the model predicts a set of event instances:
\begin{equation}
    \{a_k\}_{k=1}^{K} = \Phi(\mathcal{V}), 
    \quad 
    a_k = \left(u_k, y_k, \tau_k \right),
\end{equation}
where $u_k$ denotes the responsible player identity, 
$y_k \in \mathcal{Y}^{+}$ denotes the event category, 
and $\tau_k = [s_k,e_k]$ specifies the temporal interval containing the decisive visual evidence corresponding to $y_k$. Unlike conventional video-level classification, this formulation permits multiple events in the same clip and requires each event to be grounded to a specific player.

PlayNet solves this problem by converting the video of a play into a set of player trajectories with grounded identity. In such play, each player would be classified with an event category. 
The model first extracts trajectory-guided player and ball tokens through a video backbone, then enriches each token with global court context and relational information from other entities. Finally, a gated cross-clip aggregation module produces both a player-level event prediction and a coarse temporal evidence score over clips.

\begin{figure*}[t]
    \centering
    \includegraphics[width=\textwidth]{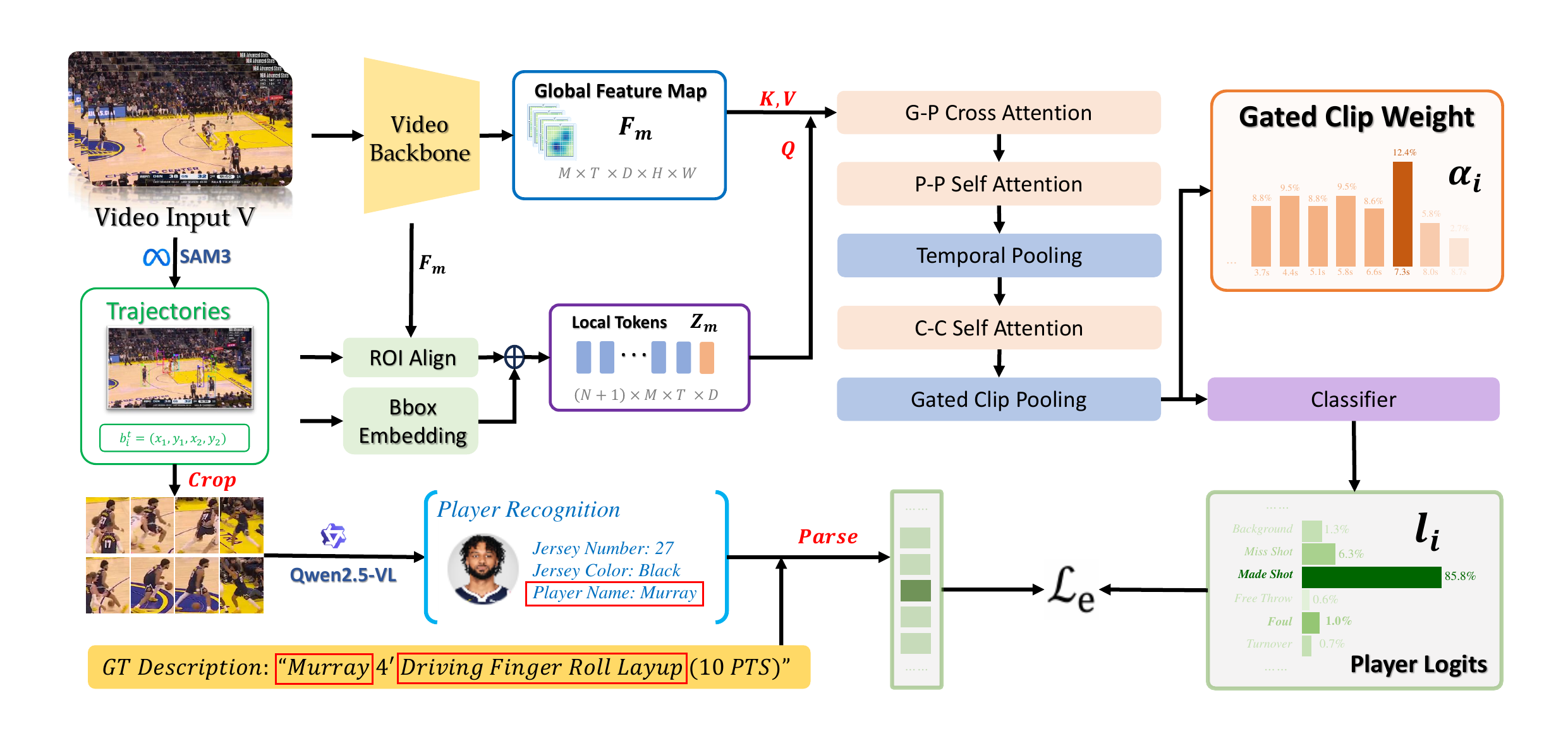}
    \vspace{-15pt}
    \caption{
        \textbf{Architecture of the proposed \model.}
        \model\, converts a basketball video into identity-grounded player and ball trajectories, extracts trajectory-guided entity tokens from dense backbone features, and models global context, player-player, and cross-clip interactions.
        Gated clip pooling produces a player-level representation for event classification, while the learned clip weights provide coarse temporal evidence for the predicted event.
    }
    \label{fig:method}
    \vspace{-9pt}
\end{figure*}

\subsection{Identity-grounded Entity Trajectories} \label{subsec:trajectory_tracking_player_recognition} 
Firstly, \model\ convert anonymous visual entities in the video into structured and identity-grounded trajectories, with each trajectory associated with an identity from real-game roster. 
PlayNet uses players and the ball as the atomic entities for reasoning. We first track all visible on-court players and the ball throughout the video. Let
\begin{equation*}
    \mathcal{B} = \{B_i\}_{i=1}^{N}, \quad 
    B_i = \{b_i^t\}_{t=1}^{T}
\end{equation*}
denote the set of $N$ player trajectories, where $b_i^t$ is the bounding box of player $i$ in frame $t$. Similarly, the ball trajectory is denoted as :
\begin{equation*}
    O = \{o^t\}_{t=1}^{T}.
\end{equation*}
We obtain these trajectories with $T\times (N+1)$ bounding boxes from the video using SAM3\cite{sam3} with textual prompts for ``basketball player on the court'' and ``basketball''. 

To associate each player trajectory with a roster identity, we uniformly sample frames from the video, crop the corresponding player regions, 
and let Qwen2.5-VL\cite{Qwen25VL} predicts jersey color and number with the game roster as context.
After that, each trajectory $B_i$ is assigned with an identity $u_i$.
Thus, each visual player token is linked to a real player identity.

\subsection{Trajectory-guided Visual Encoding}
To enable \model\ to know \emph{what} does every player do in a play, each player trajectory will be further associated with an event label.
Given the identity-grounded player trajectories and the ball trajectory, we then extract visual tokens for each player and the ball as illustrated in Fig.~\ref{fig:method}. 
We divide the input video into $M$ temporal clips $\{X_m\}_{m=1}^M$, each containing $L$ sampled frames. A TimeSformer backbone $f_{\theta}$ encodes each clip into a spatio-temporal feature map:
\begin{equation}
F_m = f_{\theta}(X_m), \quad F_m \in \mathbb{R}^{L \times H \times W \times D}, 
\end{equation}
where $D$ is the feature dimension and $H \times W$ is the spatial resolution.
For the $l^{th}$ frame in the $m^{th}$ clip, we use ROIAlign\cite{maskrcnn} to extract the feature of $i^{th}$ entity, denoted as $r_{m,i}^l$, from the backbone feature map according to the corresponding player or ball boundng box:
\begin{equation}
    r_{m,i}^l = \text{ROIAlign}(F_m^l, b_{m,i}^l) \in \mathbb{R}^D.
\end{equation}
In order to better consider the spatial and semantic information, we add two embeddings to the feature map of each entity:
(i) As for entity with bounding box $\text{b}\in\{b_{m,i}^l, b_{m,o}^l\}$, geometric embedding $\textbf{e}_{\text{geo}}=\phi_{\text{box}}(\text{b})$ encodes its normalized box coordinates, center location, size, aspect ratio, area, and temporal displacement.  
(ii) Type embedding $\textbf{e}_{\text{type}}\in\{\textbf{e}_{\text{player}},\textbf{e}_{\text{ball}}\}$ is also considered to indicate the type of entity.
Therefore, each entity feature could be represented as follows for player and ball:
\begin{equation}
\begin{aligned}
    z_{m,i}^{l}
    &= r_{m,i}^{l}
    + \phi_{\text{box}}\left(b_{m,i}^{l} + \mathbf{e}_{\text{player}}\right),\quad  i\in[1,N]\\
    z_{m,o}^{l}
    &= r_{m,o}^{l}
    + \phi_{\text{box}}\left(b_{m,o}^{l} + \mathbf{e}_{\text{ball}}\right).
\end{aligned}
\end{equation}
resulting the frame-level entity tokens $Z_m^l$ as:
\begin{equation}
    Z_m^l = [z_{m,1}^l, \dots, z_{m,N}^l, z_{m,o}^l] \in \mathbb{R}^{(N+1) \times D}.
\end{equation}

\subsection{Interaction-aware Evidence Aggregation}
\label{subsec:feature_aggregation}
After trajectory-guided visual encoding, each clip is represented by dense visual features and structured entity tokens. 
Specifically, we obtain a feature map 
$F \in \mathbb{R}^{M \times L \times H \times W \times D}$
and entity tokens 
$Z \in \mathbb{R}^{M \times L \times (N+1) \times D}$
, including $N$ player tokens and one ball token.
We then progressively aggregate these tokens into a final representation $h_i \in \mathbb{R}^{D}, i\in[1,N]$ for each player, 
which is used for player-level event classification. Meanwhile, \model\ also produces clip-level evidence weights $\alpha_{m,i}$ for temporal grounding.  
The above process is performed in 5 stages: (i) global-to-entity context attention, (ii) intra-frame entity interaction, (iii) within-clip temporal pooling, (iv) cross-clip evidence aggregation, and (v) event classification.

\vspace{3pt} \noindent \textbf{Global-to-entity context attention.}
Since local ROI tokens may miss court layout and surrounding play context, we allow each entity token to attend to the global feature map. 
Given the dense feature maps 
$F \in \mathbb{R}^{M \times L \times H \times W \times D}$, 
we flatten the spatial dimensions and obtain scene tokens $G \in \mathbb{R}^{M \times L \times HW \times D}.$
For each frame, the player and ball tokens attend to the corresponding scene tokens:
\begin{equation}
    \tilde{Z} 
    =
    \mathrm{CrossAttn}(Q = Z, K = G, V = G)\in \mathbb{R}^{M \times L \times (N+1) \times D},
\end{equation}
where the attention is applied independently for each clip and frame. 
This operation injects court-level and scene-level information into the entity tokens.

\vspace{3pt} \noindent \textbf{Intra-frame entity interaction.}
After injecting global scene context, we further model interactions among players and the ball within each frame. 
Given the context-enhanced entity tokens 
$\tilde{Z}$, 
we apply self-attention over the $N+1$ entity tokens in each frame:
\begin{equation}
    \bar{Z} = \mathrm{SelfAttn}(\tilde{Z})\in \mathbb{R}^{M \times L \times (N+1) \times D},
\end{equation}
where the attention is performed independently for each clip and frame along the entity dimension, 
allowing each player token to incorporate information from teammates, opponents, and the ball. 
Here, the ball token participates in the interaction reasoning as an auxiliary entity, while the player tokens remain the main event candidates for later classification.

\vspace{3pt} \noindent \textbf{Within-clip temporal pooling.}
Similar as before, the ball token is not directly classified as an event instance. Only player tokens are pooled into player-level clip representations.
For the $i^{th}$ player, we aggregate its relation-enhanced frame tokens $c_{m,i}$ within a clip:
\begin{equation}
    c_{m,i} = \mathrm{TemporalPool}(\{\bar{z}_{m,i}^{l}\}_{l=1}^{L}) \in\mathbb{R}^{D}.
\end{equation}

\vspace{3pt} \noindent \textbf{Cross-clip evidence aggregation.}
Furthermore, we obtain clip-level player representations $C \in \mathbb{R}^{M \times N \times D}$ by aggregating features of segmented clips,
where $M$ is the number of clips and $N$ is the number of tracked players. 
We then aggregate information across clips to capture video-level temporal context. 
This allows each player representation to incorporate evidence from different temporal segments before the final event prediction.
We first apply self-attention along the clip dimension:
\begin{equation}
    \hat{C} = \mathrm{ClipAttn}(C)\in \mathbb{R}^{M \times N \times D}.
\end{equation}
Through these features, we compute a normalized weight $\alpha_{m,i}$ over clips for each player, indicating the temporal segment of evidence, denoted as:
\begin{equation}
    \alpha_{m,i}
    =
    \frac{\exp(\psi(\hat{c}_{m,i}))}
    {\sum_{m'=1}^{M}\exp(\psi(\hat{c}_{m',i}))},
\end{equation}
where $\psi(\cdot)$ is a lightweight MLP. 
The learned weights $\alpha_{m,i}$ serve two purposes:
First, they allow the classifier to focus on several clips instead of treating all clips equally. 
Second, they provide coarse temporal grounding by indicating which clip contributes most to the predicted event for each player.
Finally, the video-level representation of the $i^{th}$ player is obtained by gated clip pooling:
\begin{equation}
    h_i
    =
    \sum_{m=1}^{M}
    \alpha_{m,i}\hat{p}_{m,i}
    \in
    \mathbb{R}^{D}.
\end{equation}
By stacking all player representations, we derive the player representation $H = [h_1,\dots,h_N] \in \mathbb{R}^{N \times D}$ with the information of other entities, frames and surrounding video clips.

\vspace{3pt} \noindent \textbf{Event classification.}
In each play, we classify the players into different event category from $\mathcal{Y}^{+}$ using the player representation $H = [h_1,\dots,h_N]$. The final representation $h_i$ of the $i^{th}$ player is fed into a shared classifier $\Phi_{\text{cls}}$:
\begin{equation}
    l_i = \Phi_{\text{cls}}(h_i), \quad \pi_i = \text{Softmax}(l_i),
\end{equation}
where $\pi_i$ is the predicted probability over ten event classes and background.

\subsection{Training and Inference}
\label{subsec:output_training}

During training, our goal is to optimize the player-level training loss $\mathcal{L}_{\mathrm{player}}$, which is a weighted cross-entropy loss with label smoothing and bootstrapped targets:
\begin{equation}
    \mathcal{L}_{\mathrm{player}}
    =
    -\frac{1}{N}\sum_{i=1}^{N}
    w_{y_i}\sum_{c\in\mathcal{Y}^{+}}
    q_{i,c}\log \pi_{i,c},
\end{equation}
where $w_{y_i}$ is the class weight and the background class is down-weighted to reduce domination by non-event players. 
Let $q_i^{\mathrm{smooth}}$ be the label-smoothed target, $\text{sg}(\cdot)$ denote stop-gradient, and $\beta$ control the contribution of the hard label,
the bootstrapped target is
\begin{equation}
    q_i = \beta q_i^{\mathrm{smooth}} + (1-\beta) \text{sg}(\pi_i).
\end{equation}
At inference time, similarly, each player trajectory is classified as 
\begin{equation}
    \hat{y}_i = \arg\max_{c\in\mathcal{Y}^{+}} \pi_{i,c}.
\end{equation}
For each player detected at beginning, the associated roster identity $u_i$ gives the `who' prediction, and $\hat{y}_i$ gives the `what' prediction. The temporal evidence is decoded from the highest-weighted clip:
\begin{equation}
    \hat{m}_i = \arg\max_{m} \alpha_{m,i}, \quad \hat{\tau}_i = [s_{\hat{m}_i}, e_{\hat{m}_i}]
\end{equation}
where $[s_m, e_m]$ is the temporal extent of clip $m$. The final output is:
\begin{equation}
    \hat{\mathcal{A}}(\mathcal{V}) = \{(u_i, \hat{y}_i, \hat{\tau}_i | \hat{y}_i \neq \text{bg}\}.
\end{equation}

\section{Experiments}
\label{sec:experiments}

This section presents the experimental evaluation of \textbf{\model}\, on \textbf{\dataset}.
We first introduce the implementation details in Sec.~\ref{sec:implementation_details}, followed by the evaluation metrics in Section~\ref{sec:eval_metric}.
We then describe the compared methods in Sec.~\ref{sec:compared_methods}, and finally, we discuss the quantitative results, ablation studies, and qualitative analysis from Sec.~\ref{sec:quantitative_results} to Sec.~\ref{sec:qualitative_results}.

\subsection{Implementation Details}
\label{sec:implementation_details}
In \model, each input video is partitioned into $M=12$ clips, with $T=8$ frames sampled per clip and resized to $224 \times 224$. 
We initialize our backbone using a TimeSformer~\cite{timesformer} model pretrained on Kinetics-400~\cite{kinetics}. 
After generating patch tokens~(patch size of $16 \times 16$), we discard the class token and reshape the remaining tokens into a $14 \times 14$ spatio-temporal feature map. 
We then apply ROIAlign~\cite{maskrcnn} with an output size of $1 \times 1$ on this feature map to extract player and ball tokens. 
We train \model for 10 epochs using the AdamW~\cite{adamw} optimizer with a learning rate of $5\times10^{-5}$, a batch size of 8, and a weight decay of 0.05.
For the training loss, we set the bootstrapping coefficient $\beta$ to 0.8 and assign the background class a weight of 0.1 in the weighted cross-entropy loss.

\subsection{Evaluation Metrics}
\label{sec:eval_metric}
We evaluate basketball event understanding on the \dataset test set from two perspectives: player-level event recognition and temporal evidence localization.

\vspace{3pt}
\noindent \textbf{Player-level event recognition.}
Ground-truth labels for each test video are represented as a set of player-event pairs.
The model takes the video as input and outputs a list of predicted player-event pairs, and a prediction is considered correct only when both the event category and the responsible player match the ground truth.
To assess retrieval performance, we report Recall@1, Recall@3, and Recall@5~(Rec@1/3/5), which measure the fraction of ground-truth pairs successfully retrieved within the top-$K$ predictions.
This evaluates the model's ability to identify active participants and their corresponding actions without being distracted by background players.
Additionally, we report the macro F1-score across event categories to evaluate recognition quality under class imbalance.

\vspace{3pt}
\noindent \textbf{Temporal evidence localization.}
As described in Sec.~\ref{subsec:temporal_subset}, we construct a temporal evaluation subset of 1,000 player-level samples, with 100 samples for each event category.
Each sample is manually annotated with the start and end timestamps of the target event.
During inference, the model produces clip-level gate weights for each player-level prediction, and we select the clip with the highest gate weight as the predicted temporal segment.

For the $i$-th sample, $[\hat{s}_i,\hat{e}_i]$ and $[s_i,e_i]$ denote the predicted and ground-truth event intervals, respectively; where $\hat{s}_i$ and $\hat{e}_i$ are the predicted start and end timestamps, while $s_i$ and $e_i$ are the ground-truth ones.
To measure temporal overlap, we employ a modified temporal IoU~(mIoU), defined as the intersection length divided by the shorter segment length:

\begin{equation*}
    \text{mIoU}_i = \frac{ \left| [\hat{s}_i,\hat{e}_i] \cap [s_i,e_i] \right| }{ \min \left( \hat{e}_i-\hat{s}_i, e_i-s_i \right) }
\end{equation*}

We further define Hit@$\theta$ to evaluate whether the model correctly localizes the temporal evidence of an event.
A sample is counted as a hit if and only if the event category is correctly predicted and the temporal overlap exceeds a threshold $\theta$:
\begin{equation*}
    \text{Hit}@\theta = \frac{1}{N_t} \sum_{i=1}^{N_t} \mathbb{I}(\hat{y}_i = y_i) \cdot \mathbb{I}(\text{mIoU}_i > \theta),
\end{equation*}
where $N_t = 1000$ is the number of annotated samples, $\hat{y}_i$ and $y_i$ denote the predicted and ground-truth event labels, and $\theta$ is the overlap threshold.
We report Hit@0.3 and Hit@0.5 to measure the localization accuracy of correctly recognized events.

\subsection{Baselines}
\label{sec:compared_methods}
We compare \model\, with three representative video classification backbones: R3D~\cite{r3d}, SlowFast~\cite{slowfast}, and TimeSformer~\cite{timesformer}.
To adapt these models to player-centric event understanding, we construct crop-based baselines using the same automatically generated player and ball tracks.
Specifically, for each player, we crop the corresponding player regions across original frames to form a player-centric video sequence, which is resized and fed into the backbone for event category prediction.
The prediction scores of all players are then used to form ranked player-event pairs, which are evaluated with the same recognition metrics as \model\,.
For R3D~\cite{r3d} and TimeSformer~\cite{timesformer}, we sample 64 frames at 4 FPS from each player-centric sequence, applying padding or truncation where necessary to maintain a uniform length.
For SlowFast~\cite{slowfast}, we sample 32 frames at 3 FPS to match its input specifications.
All compared methods are trained with the same player-level event labels.
Note that since these crop-based baselines do not produce clip-level temporal evidence, we only report temporal localization metrics for our \model.

\subsection{Quantitative Results}
\label{sec:quantitative_results}
We present quantitative comparisons on \dataset\, in Tab.~\ref{tab:quantitative_results}. 
Among the crop-based baselines, SlowFast~\cite{slowfast} achieves the highest F1-score and Rec@1, whereas TimeSformer~\cite{timesformer} excels on Rec@3 and Rec@5.
While these results indicate that isolated crops provide useful local cues, their performance remains constrained when events inherently depend on global court context, player-player relations, and player-ball interactions.
In contrast, \model\, consistently outperforms all baselines across all recognition metrics, boosting F1-score from 0.46 to 0.68 and Rec@1 from 0.45 to 0.70 over the strongest competitor.
These gains underscore the necessity of modeling the global play structure to accurately identify both the responsible player and the corresponding event.
Furthermore, \model\, achieves 0.42 Hit@0.3 and 0.38 Hit@0.5 on the temporal subset, demonstrating that its clip-level gates successfully capture meaningful, coarse-grained temporal evidence for event localization.

\begin{table*}[t]
    \centering
    \small
    \renewcommand{\arraystretch}{1.1}
    \caption{
        \textbf{Quantitative results on the test set of our \dataset.}
        Rec@1/3/5 and F1-score are macro-averaged over event categories.
        Hit@0.3/0.5 evaluate coarse temporal evidence localization.
        The best results are marked in \textcolor{red}{\textbf{red}}, and the second-best results are \underline{underlined}.
    }
    \label{tab:quantitative_results}
    \vspace{6pt}
    \resizebox{=\textwidth}{!}{
    \begin{tabular}{lcccccc}
        \toprule
        \textbf{Method} & \textbf{F1-score} & \textbf{Rec@1} & \textbf{Rec@3} & \textbf{Rec@5} & \textbf{Hit@0.3} & \textbf{Hit@0.5} \\
        \midrule
        R3D 
        & 0.412 & 0.386 & 0.717 & 0.870 & -- & -- \\

        SlowFast 
        & \underline{0.456} & \underline{0.445} 
        & 0.746 & 0.901 & -- & -- \\

        TimeSformer 
        & 0.329 & 0.322 & \underline{0.791} & \underline{0.918} & -- & -- \\

        \midrule
        \rowcolor{oursblue}
        \textbf{\model} 
        & \textcolor{red}{\textbf{0.682}} 
        & \textcolor{red}{\textbf{0.701}} 
        & \textcolor{red}{\textbf{0.890}} 
        & \textcolor{red}{\textbf{0.954}} 
        & \textcolor{red}{\textbf{0.424}} 
        & \textcolor{red}{\textbf{0.380}} \\
        \bottomrule
    \end{tabular}
    }
    \vspace{-6pt}
\end{table*}

        
        
        

\subsection{Ablation Studies}
\label{sec:ablation}
We conduct ablation studies from two perspectives to analyze \model. 
First, we perform component-wise ablations to evaluate the individual contributions of global context, player-player interactions, cross-clip self-attention, and ball dynamics. 
Second, we investigate the impact of temporal resolution and coverage by varying the sampling configuration, specifically the frame rate and the number of temporal clips.

\vspace{3pt}
\noindent\textbf{Component ablation.}
As shown in Tab.~\ref{tab:ablation}, our full model achieves the best recognition performance, yielding 0.68 F1-score, 0.70 Rec@1, 0.89 Rec@3, and 0.95 Rec@5.
Compared to the baseline~(ROI+Gate), which directly aggregates ROI features without interaction reasoning, \model\, improves F1-score from 0.52 to 0.68 and Rec@1 from 0.52 to 0.70. 
This indicates that player-centric event understanding requires structured reasoning over the entire play rather than relying solely on local player appearance.
Furthermore, removing player-player interactions or ball dynamics leads to significant performance drops, confirming that basketball events are inherently defined by synergistic interactions among players and the ball.
Excluding global context also degrades performance, suggesting that court-level scene context provides complementary cues.
Interestingly, for temporal localization, the trend slightly diverges from recognition: the ``w/o Clip SA'' variant obtains the highest Hit@0.3, whereas the full model dominates recognition metrics.
This discrepancy suggests that event classification and temporal evidence localization represent related but distinct objectives.

\begin{table*}[t]
    \centering
    \small
    \renewcommand{\arraystretch}{1.10}
    \caption{
        \textbf{Ablation study of different components in \model.}
        Rec denotes recall.
        The best results are marked in \textcolor{red}{\textbf{red}}, and the second-best results are \underline{underlined}.
    }
    \label{tab:ablation}
    \vspace{6pt}
    \resizebox{=\textwidth}{!}{
    \begin{tabular}{lccccccccc}
    \toprule
    \textbf{Method} 
    & \textbf{Global} 
    & \textbf{P-P} 
    & \textbf{Clip SA} 
    & \textbf{Ball} 
    & \textbf{F1-score} 
    & \textbf{Rec@1} 
    & \textbf{Rec@3} 
    & \textbf{Rec@5} 
    & \textbf{Hit@0.3} \\
    \midrule
    ROI+Gate 
    & -- & -- & -- & -- 
    & 0.520 & 0.522 & 0.812 & 0.932 & \underline{0.474} \\
    
    w/o Global 
    & -- & \checkmark & \checkmark & \checkmark 
    & \underline{0.662} 
    & \underline{0.673} 
    & \underline{0.875} 
    & \underline{0.952} 
    & 0.451 \\
    
    w/o P-P 
    & \checkmark & -- & \checkmark & \checkmark 
    & 0.560 & 0.565 & 0.824 & 0.941 & 0.363 \\
    
    w/o Clip SA 
    & \checkmark & \checkmark & -- & \checkmark 
    & 0.601 & 0.620 & 0.860 & 0.951 
    & \textcolor{red}{\textbf{0.588}} \\
    
    w/o Ball 
    & \checkmark & \checkmark & \checkmark & -- 
    & 0.558 & 0.571 & 0.831 & 0.939 & 0.323 \\
    
    \midrule
    \rowcolor{oursblue}
    \textbf{\model} 
    & \checkmark & \checkmark & \checkmark & \checkmark 
    & \textcolor{red}{\textbf{0.682}} 
    & \textcolor{red}{\textbf{0.701}} 
    & \textcolor{red}{\textbf{0.890}} 
    & \textcolor{red}{\textbf{0.954}} 
    & \textbf{0.424} \\
    \bottomrule
    \end{tabular}
    }
\end{table*}

    
    
    
    

\begin{table*}[t]
    \centering
    \small
    \setlength{\tabcolsep}{4.8pt}
    \renewcommand{\arraystretch}{1.12}
    \caption{
    \textbf{Hyperparameter ablation of \model.}
    ``Frames'' denotes the number of sampled frames per clip.
    The best results are marked in \textcolor{red}{\textbf{red}}, and the second-best are \underline{underlined}.
    }
    \label{tab:hyper_ablation}
    \vspace{5pt}
    \resizebox{\textwidth}{!}{
    \begin{tabular}{lcccccccc}
    \toprule
    \textbf{Setting} & \textbf{FPS} & $M$ & \textbf{Frames} 
    & \textbf{F1-score} & \textbf{Rec@1} & \textbf{Rec@3} & \textbf{Rec@5} & \textbf{Hit@0.3} \\
    \midrule
    
    \multicolumn{9}{c}{\textit{FPS Ablation: fixing $M=12$ and clip duration as 2s}} \\
    \midrule
    FPS=1 
    & 1 & 12 & 2 
    & 0.611 & 0.653 & 0.870 & 0.944 & \textcolor{red}{\textbf{0.537}} \\

    FPS=2 
    & 2 & 12 & 4 
    & 0.543 & 0.554 & 0.813 & 0.923 & 0.289 \\

    \rowcolor{oursblue}
    FPS=4 
    & 4 & 12 & 8 
    & \textcolor{red}{\textbf{0.682}} 
    & \textcolor{red}{\textbf{0.701}} 
    & \textcolor{red}{\textbf{0.890}} 
    & \underline{0.954}
    & \underline{0.424} \\
    
    \rowcolor{secondblue}
    FPS=8 
    & 8 & 12 & 16 
    & \underline{0.642} 
    & \underline{0.657}
    & \underline{0.887} 
    & \textcolor{red}{\textbf{0.958}}
    & 0.394 \\
    
    \midrule
    \multicolumn{9}{c}{\textit{Clip Number Ablation: fixing FPS=4 and clip duration as 2s}} \\
    \midrule
    $M=4$ 
    & 4 & 4 & 8 
    & 0.513 & 0.517 & 0.760 & 0.895 & 0.222 \\

    $M=8$ 
    & 4 & 8 & 8 
    & 0.577 & 0.605 & 0.836 & 0.928 & 0.384 \\
    
    \rowcolor{oursblue}
    $M=12$ 
    & 4 & 12 & 8 
    & \textcolor{red}{\textbf{0.682}} 
    & \textcolor{red}{\textbf{0.701}} 
    & \textcolor{red}{\textbf{0.890}} 
    & \textcolor{red}{\textbf{0.954}} 
    & \textcolor{red}{\textbf{0.424}} \\
    
    \rowcolor{secondblue}
    $M=16$ 
    & 4 & 16 & 8 
    & \underline{0.637} 
    & \underline{0.654} 
    & \underline{0.882} 
    & \underline{0.951}
    & \underline{0.421} \\
    \bottomrule
    \end{tabular}
    }
    \vspace{-9pt}
\end{table*}

\vspace{3pt}
\noindent\textbf{Sampling ablation.}
As presented in Tab.~\ref{tab:hyper_ablation}, we investigate the impact of sampling frame rate~(FPS) and the number of clips $M$.
With $M=12$ fixed, FPS=4 yields the best overall recognition performance~(0.68 F1-score and 0.70 Rec@1).
Lower frame rates suffer from insufficient motion details, whereas a higher rate~(FPS=8) offers marginal gains on high-recall metrics at the cost of a slightly lower F1-score. 
This suggests that FPS=4 optimally balances temporal resolution and frame redundancy.
On the other hand, when fixing FPS=4, increasing $M$ from 4 to 12 consistently boosts performance, demonstrating that a broader temporal span helps the model capture sparse event evidence.
However, further increasing $M$ to 16 leads to a little lower performance across all metrics compared with $M=12$. This suggests that after the full video has been covered, additional clips mainly introduce redundant or highly overlapping temporal context, which may distract top-ranked player-event predictions. 
Consequently, FPS = 4 and $M=12$ strike the best trade-off between recognition accuracy and temporal localization.

\subsection{Qualitative Results}
\label{sec:qualitative_results}

\begin{figure*}[t]
    \centering
    \includegraphics[width=\textwidth]{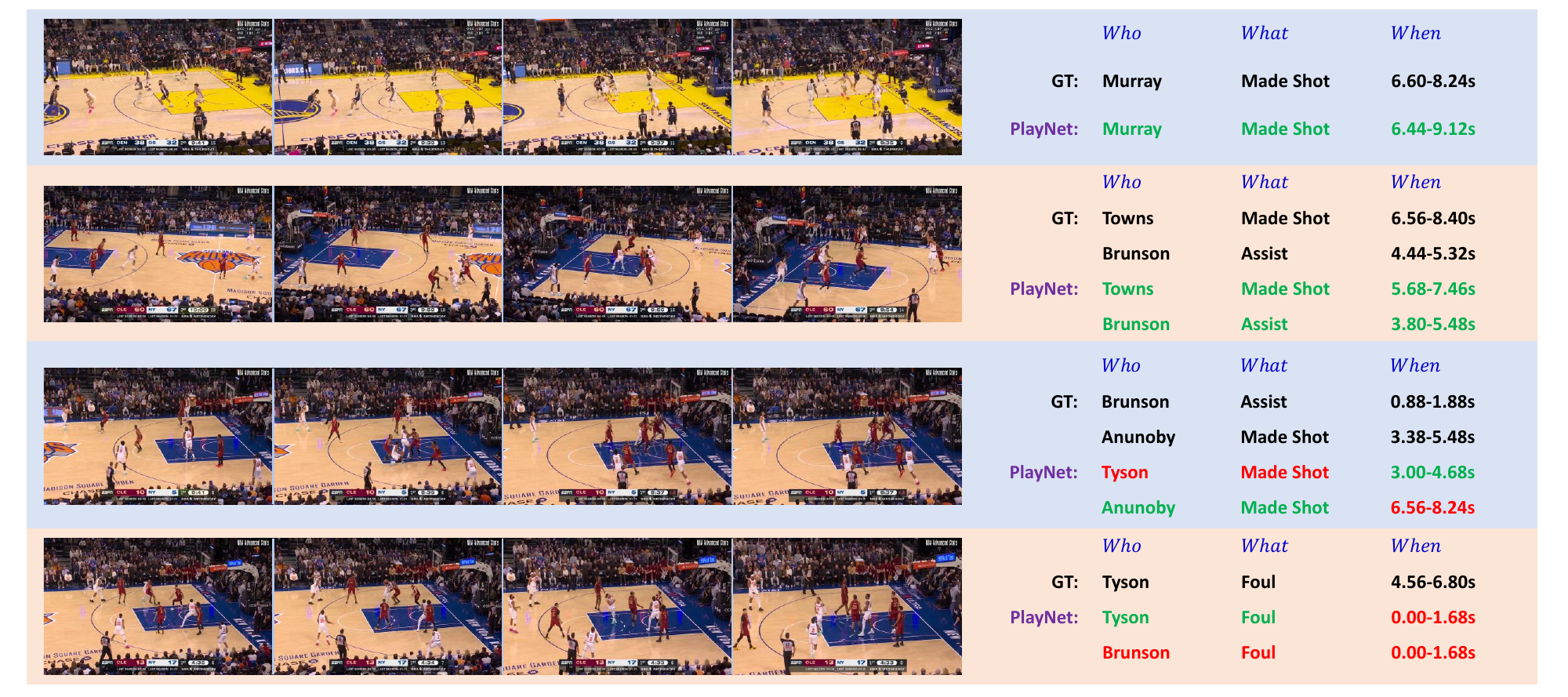}
    \vspace{-8pt}
    \caption{
    \textbf{Qualitative results of \model.}
    Each example shows sampled video frames and the corresponding player-grounded event predictions in terms of \emph{who}, \emph{what}, and \emph{when}.
    Ground-truth annotations are shown in black.
    For PlayNet predictions, correct predictions are highlighted in green, while incorrect predictions are highlighted in red.
    }
    \label{fig:qualitative_results}
    \vspace{-8pt}
\end{figure*}

Fig.~\ref{fig:qualitative_results} shows qualitative examples of \model{} on \dataset.
Each example contains sampled video frames and the corresponding player-grounded event predictions.
The results demonstrate that \model{} can identify the responsible player, recognize the event category, and provide coarse temporal evidence through gated clip weights.
In the first two examples, \model{} correctly predicts both single-event and multi-event cases, including the responsible players, event categories, and approximate temporal segments.
The last two examples illustrate failure cases, where the model may confuse the responsible player or localize the event to an incorrect temporal segment.
These cases suggest that player-centric basketball understanding remains challenging when multiple players are spatially close or when the discriminative event evidence is temporally brief.

\section{Conclusion}
\label{sec:conclusion}
In this paper, we focus on player-centric basketball event understanding, a challenging task that moves beyond conventional video-level classification to resolve ``who did what and when''.
To support this task, we introduce \dataset, a player-centric basketball event understanding dataset that grounds event labels to responsible players and provides a manually annotated subset for evaluating temporal evidence localization. 
Based on this data, we propose \model, a player-centric reasoning framework that tracks players and the ball, associates visual trajectories with player identities, and predicts player-level events by modeling global court context, player-player interactions, player-ball relations, and sparse temporal evidence across clips.
Our experiments demonstrate that \model significantly outperforms representative video-level and crop-based baselines; furthermore, temporal evaluations confirm that our clip-level gates successfully capture meaningful, coarse-grained evidence for event localization. 
Overall, our task formulation, benchmark, and framework lay a solid foundation for structured, fine-grained basketball video understanding.

\bibliography{egbib}
\end{document}